\documentclass{article}

\usepackage{PRIMEarxiv}
\usepackage{natbib}
\usepackage{CJKutf8}
\usepackage{CJK}
\usepackage[utf8]{inputenc} 
\usepackage[T1]{fontenc}    
\usepackage{hyperref}       
\usepackage{url}            
\usepackage{booktabs}       
\usepackage{amsfonts}       
\usepackage{nicefrac}       
\usepackage{microtype}      
\usepackage{lipsum}
\usepackage{graphicx}
\usepackage{amsmath}
\usepackage{pifont}
\usepackage[inline]{enumitem}
\usepackage{enumitem}
\usepackage{boldline}
\usepackage{hhline}
\usepackage{multirow}
\usepackage{xcolor}
\usepackage{fontawesome5}
\usepackage{tablefootnote} 
\usepackage{listings}
\usepackage{graphicx}
\usepackage{tabularx}
\usepackage{xspace} 

\usepackage[utf8]{inputenc}
\graphicspath{{images/}}     
\usepackage{amsthm}
\theoremstyle{definition}

\usepackage[most]{tcolorbox}

\usepackage[most]{tcolorbox}
\usepackage{verbatim} 
\usepackage{booktabs}
\usepackage{multirow}
\usepackage{makecell}
\usepackage{siunitx}
\usepackage{wrapfig}

\usepackage{threeparttable}
\usepackage[table]{xcolor}

\newcommand{\best}[1]{\textbf{#1}}
\newcommand{\second}[1]{\underline{#1}}

\usepackage[most]{tcolorbox}
\usepackage{xcolor}

\definecolor{casebg}{RGB}{248,249,251}
\definecolor{caseframe}{RGB}{214,220,229}

\newtcolorbox{casebox}[1]{
  enhanced,
  breakable,
  colback=casebg,
  colframe=caseframe,
  boxrule=0.45pt,
  arc=1.2mm,
  left=1.4mm,
  right=1.4mm,
  top=1.0mm,
  bottom=1.0mm,
  title={#1},
  fonttitle=\bfseries,
  coltitle=black,
}

\newtcolorbox{findingbox}{
  colback=black!5!white,      
  colframe=black!75!black,    
  arc=3mm,                    
  boxrule=0.5pt,              
  left=4mm, right=4mm, top=3mm, bottom=3mm 
}
\usepackage[most]{tcolorbox}

\usepackage[most]{tcolorbox}

\tcbset{
  prompt_box1_style/.style={
    enhanced,
    attach boxed title to top left={
        xshift=-2mm,
        yshift=-2mm
    },
    boxed title style={
        rounded corners,
        size=small,
        colback=gray!20
    },
    coltitle=black,
    colback=white,
    colframe=black,
    boxrule=0.5pt,
    sharp corners, 
    colbacktitle=gray!20,
    coltitle=black,
    rounded corners,
    segmentation style={draw=none},
    sharp corners=north, 
    boxrule=0.5pt,
    drop shadow=black!50!white,
    left=4mm, right=4mm, top=4mm, bottom=4mm, 
    fontupper=\small,
    boxed title style={
        rounded corners,
        size=small,
        colback=gray!20
    }
  }
}

\tcbset{
  prompt_box2_style/.style={
    enhanced,
    boxed title style={
        rounded corners,
        size=small,
        colback=gray!20
    },
    coltitle=black,
    colback=white,
    colframe=black,
    boxrule=0.5pt,
    sharp corners, 
    colbacktitle=gray!20,
    coltitle=black,
    rounded corners,
    segmentation style={draw=none},
    sharp corners=north, 
    boxrule=0.5pt,
    drop shadow=black!50!white,
    fontupper=\fontsize{9pt}{8pt}\selectfont,
    boxed title style={
        rounded corners,
        size=small,
        colback=gray!20
    }
  }
}

\newtcolorbox{caselogbox}[1]{
  prompt_box1_style, 
  title=#1          
}

\newtcolorbox{promptlogbox}[1]{
  prompt_box2_style, 
  title=#1          
}

\usepackage{tcolorbox}
\usepackage{listings} 
\tcbuselibrary{breakable, listings} 

\tcbuselibrary{listings} 
\usepackage[most]{tcolorbox}

\newtcolorbox{textBox}[1][]{
  colback=gray!5,
  colframe=gray!60!black,
  fonttitle=\bfseries,
  colbacktitle=gray!85!black,
  title=#1,
  breakable
}

\newtcblisting{promptbox}[1][]{
  colback=gray!5,
  colframe=gray!60!black,
  fonttitle=\bfseries,
  colbacktitle=gray!85!black,
  title=#1,
  breakable,
  listing only,
  listing options={
    basicstyle=\ttfamily\small,
    breaklines=true,
    columns=fullflexible
  }
}

\title{PsychAgent: An Experience-Driven Lifelong Learning Agent for Self-Evolving Psychological Counselor}

\author{
  \textbf{Yutao Yang}$^{1}$, \textbf{Junsong Li}$^1$, Qianjun Pan$^{1}$, Jie Zhou$^{1,2}$\footnotemark[1], \textbf{Kai Chen}$^2$, \textbf{Qin Chen}$^1$, \textbf{Jingyuan Zhao}$^1$, \\ \textbf{Ningning Zhou}$^3$, \textbf{Xin Li}$^2$, \textbf{Liang He}$^1$ \\
  $^1$ School of Computer Science and Technology, East China Normal University, \\
   $^2$ Shanghai AI Laboratory, $^3$ School of Psychology and Cognitive Science, East China Normal University, \\ 
  \texttt{\{jzhou, qchen, lhe\}@cs.ecnu.edu.cn}, \\
  \textcolor{red}{\url{https://github.com/ECNU-ICALK/PsychAgent}} 
}

\begin{document}

\renewcommand{\thefootnote}{\fnsymbol{footnote}}

\maketitle

\footnotetext[1]{Corresponding Authors}

\begin{abstract}
Existing methods for AI psychological counselors predominantly rely on supervised fine-tuning using static dialogue datasets. However, this contrasts with human experts, who continuously refine their proficiency through clinical practice and accumulated experience. To bridge this gap, we propose an Experience-Driven Lifelong Learning Agent (\texttt{PsychAgent}) for psychological counseling.
First, we establish a Memory-Augmented Planning Engine tailored for longitudinal multi-session interactions, which ensures therapeutic continuity through persistent memory and strategic planning. 
Second, to support self-evolution, we design a Skill Evolution Engine that extracts new practice-grounded skills from historical counseling trajectories. Finally, we introduce a Reinforced Internalization Engine that integrates the evolved skills into the model via rejection fine-tuning, aiming to improve performance across diverse scenarios.
Comparative analysis shows that our approach achieves higher scores than strong general LLMs (e.g., GPT-5.4, Gemini-3) and domain-specific baselines across all reported evaluation dimensions. These results suggest that lifelong learning can improve the consistency and overall quality of multi-session counseling responses.
\end{abstract}


\section{Introduction}
Psychological counseling remains difficult to access in many settings because demand for mental-health support exceeds the supply of qualified professionals. Recent large language models (LLMs), including foundation models such as GPT-4o~\citep{hurst2024gpt}, Llama~\citep{touvron2023llama}, and DeepSeek~\citep{liu2024deepseek}, have shown strong semantic understanding, making them promising for tasks ranging from empathetic support to preliminary psychological assistance.

In response, the research community has increasingly specialized LLMs for mental health. Early efforts, such as ChatCounselor~\citep{liu2023chatcounselor} and SoulChat~\citep{chen2023soulchat}, used domain-specific fine-tuning to improve professional capability and emotional resonance. More recent work has moved toward agent-based systems with stronger clinical structure. For example, AutoCBT~\citep{xu2025autocbt} employs multi-agent collaboration to deliver structured Cognitive Behavioral Therapy, while PsyDT~\citep{xie2025psydt} introduces “Digital Twins” to model the personalized styles of specific therapists. Addressing the temporal dimension, MusPsy~\citep{wang2025psychological} emphasizes the importance of multi-session tracking. However, many of these systems remain largely static after training: whether relying on fixed weights or pre-defined workflows, they provide limited support for updating strategies from ongoing interactions.

This static paradigm stands in stark contrast to the professional growth of human counselors. A human therapist does not become an expert solely by studying textbooks or datasets; rather, they cultivate expertise through an \textbf{experience-driven loop} by accumulating ``clinical intuition" from years of successful and unsuccessful cases. 
This discrepancy suggests three key challenges in current AI counseling: 1) unlike human experts who refine techniques and abstract practical skills from experience, many SFT-based models remain fixed after training and do not update their strategies from interaction history; 2) counseling is inherently longitudinal, and human counselors rely on long-term memory and strategic planning to guide clients over weeks or months, whereas current models often lack persistent context for multi-session coherence; 3) human experts gradually transform accumulated knowledge into more internalized abilities that can transfer across scenarios, while existing models often depend on external retrieval or rigid prompts, with limited evidence that these learned behaviors are reflected more directly in model parameters.

To address this gap, we propose an Experience-Driven Lifelong Learning Agent (\texttt{PsychAgent}) for AI psychological counseling. Drawing inspiration from how human counselors improve through practice, our framework combines a Memory-Augmented Planning Engine for longitudinal continuity, a Skill Evolution Engine for extracting practice-grounded therapeutic skills from historical counseling trajectories, and a Reinforced Internalization Engine for integrating these evolved skills into model parameters via rejection fine-tuning. Experimental results show that PsychAgent achieves higher scores than strong general LLMs (e.g., GPT-5.4~\citep{openai2026gpt54}, Gemini-3~\citep{google2025gemini3} and domain-specific baselines, supporting the usefulness of learning from accumulated counseling experience.

Our main contributions are summarized as follows:
\begin{itemize}[leftmargin=*, align=left] 
    \item We propose \texttt{PsychAgent}, which, to our knowledge, is among the first frameworks to formulate AI psychological counseling as an experience-driven lifelong learning problem, loosely inspired by the professional development of human counselors.
    \item We design a Memory-Augmented Planning Engine for longitudinal continuity, and a Skill Evolution mechanism combined with Reinforced Internalization to enable the autonomous discovery and assimilation of therapeutic skills. Extensive experiments show that our approach achieves state-of-the-art performance, demonstrating superior generalization and consistency in providing professional counseling services compared to existing methods.
    \item Extensive experiments show that our approach achieves state-of-the-art performance, demonstrating superior generalization and consistency in providing professional counseling services compared to existing methods.
\end{itemize}

\section{Related Work}

\subsection{LLMs for Psychological Counseling}

Recent work has adapted LLMs to mental-health support and psychological counseling through domain-specific data and supervision. Early efforts such as ChatCounselor~\citep{liu2023chatcounselor}, Psy-LLM~\citep{lai2023psy}, SoulChat~\citep{chen2023soulchat}, and MindChat~\citep{MindChat} mainly improve empathy, emotional support, and counseling-oriented dialogue generation. Subsequent work injects stronger therapeutic structure into the model or system design, for example through cognitive reframing~\citep{xiao2024healme}, CBT-grounded modeling~\citep{lee2024cactus,na2024cbt}, multi-agent coordination~\citep{xu2025autocbt}, or therapist-style modeling via digital twins~\citep{xie2025psydt}. In parallel, evaluation has become more systematic, covering counseling competency, therapy-specific ability, and ethical risks~\citep{nguyen2025large,zhang2025cbt,iftikhar2025llm}, together with broader human-evaluation frameworks for healthcare LLMs~\citep{tam2024framework}.

A parallel line of work emphasizes that counseling is inherently longitudinal rather than single-turn. MusPsy~\citep{wang2025psychological}, PsyDial~\citep{qiu2025psydial}, and PsychEval~\citep{pan2026psycheval} move toward multi-session data and evaluation, while TheraMind~\citep{hu2026theramindstrategicadaptiveagent} incorporates cross-session planning and adaptation into the agent itself. However, these systems are still largely static after training: they may better model history or follow structured workflows, but they do not explicitly extract new therapeutic skills from accumulated interactions and internalize them into model parameters. Our work targets this gap by framing AI psychological counseling as an experience-driven lifelong learning problem.

\subsection{Experience-Driven Lifelong Learning}

The evolution of Continual Learning (CL) has undergone a paradigm shift, transitioning from merely mitigating catastrophic forgetting in static architectures to enabling dynamic adaptation in LLMs \citep{yang2025recent}. Traditional approaches largely addressed the stability-plasticity dilemma through regularization techniques like EWC~\citep{kirkpatrick2017overcoming} or experience replay mechanisms~\citep{rolnick2019experience, hadsell2020embracing}. While recent frameworks have introduced sophisticated memory management~\citep{huai2025task} and reinforcement learning~\citep{yang2025reinforced} to handle long-term interactions, they predominantly focus on data retention rather than high-level skill acquisition, often struggling with the semantic complexity of open-ended real-world tasks~\citep{wang2024comprehensive}.


Building on this direction, recent work has increasingly explored Self-Evolving Agents that refine capabilities through recursive loops of experience and reflection~\citep{gao2025survey, zheng2025lifelong}. Frameworks such as SAGE~\citep{liang2025sage} and Richelieu~\citep{guan2024richelieu} suggest that agents can improve reasoning through self-reflective critique. However, many existing systems still rely heavily on transient in-context learning or external retrieval (RAG) to exhibit “growth.” As formalized by Cai et al.~\citep{cai2025building}, Experience-Driven Lifelong Learning (ELL) ideally requires agents to autonomously acquire, refine, and use skills from dynamic environments. In this work, we apply this perspective to psychological counseling by modeling a cycle of accumulating clinical experience, abstracting therapeutic skills, and internalizing them through training.

\begin{figure}[t!]
    \centering
    \includegraphics[width=0.99\textwidth]{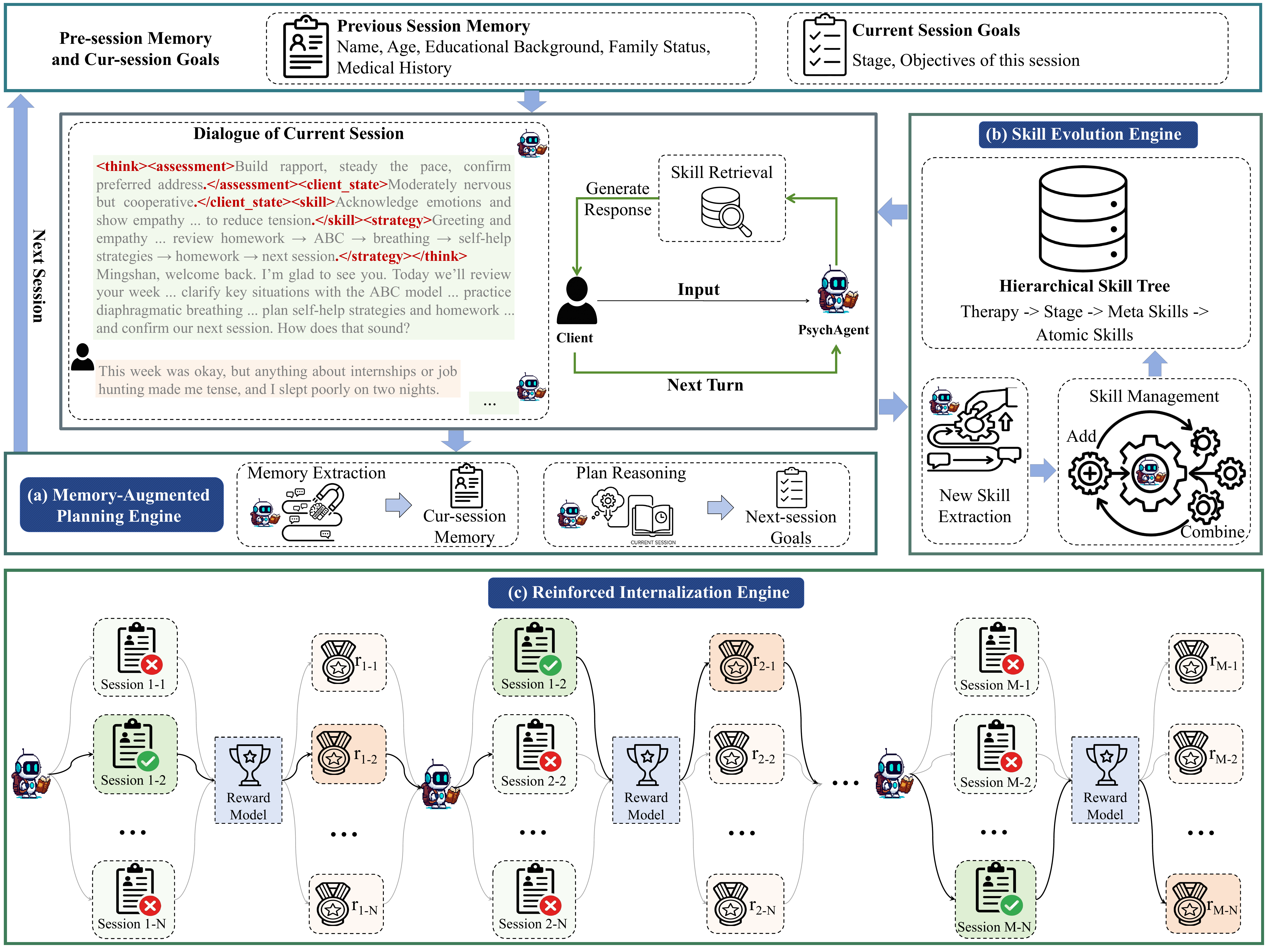}
    \vspace{-2mm}
    \caption{The overall architecture of \texttt{PsychAgent}, an experience-driven lifelong learning framework for psychological counseling. The system operates through a synergistic closed-loop mechanism: (a) The Memory-Augmented Planning Engine ensures therapeutic continuity across longitudinal sessions by reasoning over dynamic client profiles and formulating strategic goals; (b) The Skill Evolution Engine manages a hierarchical repository of therapeutic techniques, supporting both real-time context-aware retrieval and the post-session abstraction of novel skills; (c) The Reinforced Internalization Engine solidifies these capabilities via rejection fine-tuning, where optimal trajectories are selected to transform explicit skills into the model's endogenous intuition.}
    \label{fig:framework}
\vspace{-2mm}
\end{figure}

\section{Our Methods}
As illustrated in Figure~\ref{fig:framework}, we propose \texttt{PsychAgent}, an experience-driven framework intended to capture aspects of how counselors improve through practice. The architecture includes three components: the Memory-Augmented Planning Engine, which supports therapeutic continuity across longitudinal sessions through long-term context; the Skill Evolution Engine, which extracts successful intervention strategies into a structured hierarchical skill tree; and the Reinforced Internalization Engine, which incorporates these explicit skills into model parameters via rejection-based reinforcement. By iterating over these stages, \texttt{PsychAgent} is designed to reduce reliance on static imitation and heavy external retrieval.

We formulate the AI psychological counseling task as a Lifelong Sequential Decision-Making problem, where an agent $\pi_\theta$ interacts across a longitudinal timeline $\mathcal{T} = \{S_1, S_2, \dots, S_T\}$. For each session $S_t$, the agent generates a response trajectory conditioned on the evolving client state $\mathcal{M}_t$ (Memory) and a dynamic repository of therapeutic strategies $\mathcal{K}_t$ (Skills). The overarching objective is to maximize the cumulative therapeutic effectiveness $\sum_{t=1}^T \mathcal{R}(S_t)$ through an iterative closed-loop process: \textit{intra-session}, the agent utilizes a Memory-Augmented Planning Engine and a Skill Evolution Engine to retrieve explicit knowledge for immediate reasoning; \textit{inter-session}, the agent employs a Reinforced Internalization Engine to lock in the optimal trajectory $S_t^*$---identified via rejection sampling from parallel rollouts---thereby continuously updating its parameters $\theta$ to internalize successful clinical experiences.

\subsection{Memory-Augmented Planning Engine}
\label{sec:mape}

Standard LLMs often suffer from temporal incoherence in long-context interactions, failing to maintain a consistent therapeutic trajectory. To mitigate this, the Memory-Augmented Planning Engine (Figure~\ref{fig:framework}a) serves as the agent's cognitive substrate, ensuring longitudinal continuity. 
This engine is designed to support longitudinal continuity through three functions: \textit{Dynamic Memory Construction}, \textit{Strategic Planning}, and \textit{Context-Aware Generation}.

\paragraph{Dynamic Memory Construction.}
Unlike static context windows, we maintain a structured memory store $\mathcal{M}_t$ that evolves alongside the therapy. This is composed of two parts:
\begin{itemize}[leftmargin=*, align=left] 
    \item \textbf{Evolving Client Profile ($\mathcal{P}_t$):} Rather than a fixed intake form, the client profile is treated as a dynamic semantic memory. At the conclusion of each session $S_{t-1}$, the agent explicitly extracts new attributes (e.g., changes in family status, new medical diagnoses) to incrementally refine the profile:
    \begin{equation}
        \mathcal{P}_t \leftarrow \text{Update}(\mathcal{P}_{t-1}, \text{Extract}_{\text{attr}}(S_{t-1}))
    \end{equation}
    This helps the agent maintain an updated understanding of the client’s life circumstances.
    \item \textbf{Episodic Session Summaries ($\mathcal{H}_t$):} To compress long-term history, the agent generates a structured summary $h_{t-1}$ of the previous interaction, capturing key emotional shifts and intervention outcomes, which is appended to the history: $\mathcal{H}_t = \mathcal{H}_{t-1} \cup \{h_{t-1}\}$.
\end{itemize}

\paragraph{Goal-Oriented Strategic Planning.}
Therapy requires foresight. Before initiating session $S_t$, the engine performs \textit{Plan Reasoning} to synthesize the updated memory $\mathcal{M}_t = (\mathcal{P}_t, \mathcal{H}_t)$ into a coherent roadmap. The model predicts a structured plan $G_t$, encompassing:
1) Therapeutic Stage: Identifying the global phase (e.g., \textit{Case Conceptualization}, \textit{ Core Intervention}, \textit{Consolidation and Prevention}); 2) Session Objectives: Formulating specific, actionable goals for the current session (e.g., ``Reinforce cognitive restructuring techniques"). 
This planning step provides a more structured framework for subsequent dialogue generation, rather than leaving the interaction purely reactive.

To endow the agent with these capabilities, we construct high-quality instruction data $\mathcal{D} = \{(x, y)\}$ for each sub-task and train the model to minimize the negative log-likelihood across all tasks:
\begin{equation}
    \mathcal{L}_{\text{SFT}} = - \sum_{(x, y) \in \mathcal{D}_{\text{mem}} \cup \mathcal{D}_{\text{plan}} \cup \mathcal{D}_{\text{resp}}} \log P_\theta(y | x)
\end{equation}
Specifically, the training corpus is constructed to cover the core stages of the counselor’s cognitive workflow, mapping distinct inputs $x$ to targeted outputs $y$ across three tasks: for Memory Extraction (($\mathcal{D}_{\text{mem}}$)), the model learns to distill raw dialogue history into structured profile updates and session summaries; for Plan Reasoning ($\mathcal{D}_{\text{plan}}$), it learns to translate the accumulated memory state $\mathcal{M}_t$ into strategic goals $G_t$; and for Response Generation ($\mathcal{D}_{\text{resp}}$), it conditions on memory, plan, and user input ($(u_i, \mathcal{M}_t, G_t)$) to jointly generate chain-of-thought reasoning $z_i$ and the final therapeutic response $r_i$. Explicit supervision of these steps is intended to improve the model’s structured counseling decision-making.

\subsection{Skill Evolution Engine}
\label{sec:see}

While the Memory Engine anchors the agent in the client’s past, the Skill Evolution Engine (Figure~\ref{fig:framework}b) is designed to expand the model’s explicit repository of therapeutic techniques over time. Unlike static systems relying on fixed prompt engineering, this engine maintains a dynamic ontology of therapeutic techniques. We unify both the \textit{utilization} and \textit{evolution} of skills under a generative framework, acquired through rigorous Supervised Fine-Tuning (SFT) on expert-annotated cognitive tasks.

\paragraph{Hierarchical Skill Taxonomy.}
We formalize the skill knowledge base as a hierarchical tree structure $\mathcal{T} = (\mathcal{N}, \mathcal{E})$, where nodes $\mathcal{N}$ represent therapeutic skills organized into four granular levels:
\begin{itemize}[leftmargin=*, align=left]
    \item \textbf{Root Paradigms ($n_{root}$):} The foundational counseling frameworks that define the theoretical basis (e.g., CBT, Humanistic Therapy).
    \item \textbf{Therapeutic Stages ($n_{stage}$):} The macro-phases of the counseling process, specifically defined as \textit{Case Conceptualization}, \textit{Core Intervention}, and \textit{Consolidation and Prevention}.
    \item \textbf{Meta Skills ($n_{meta}$):} High-level strategic categories nested within each stage (e.g., \textit{Empathy Building} in the conceptualization phase).
    \item \textbf{Atomic Skills ($n_{atomic}$):} Concrete, executable intervention tactics used in single turns (e.g., ``Socratic Questioning," ``Metaphorical Reflection").
\end{itemize}

\paragraph{Context-Aware Skill Retrieval.}
During inference, the agent must select the optimal intervention strategy $\kappa^*$ from the skill tree $\mathcal{T}_t$ to guide the current response. This is modeled as a context-dependent generation task. Given the current dialogue state $s_t = (u_i, \mathcal{M}_t, G_t)$, the model computes the probability of selecting specific atomic skills:
\begin{equation}
    \kappa^* = \operatorname*{argmax}_{\kappa \in \mathcal{N}_{atomic}} P_\theta(\kappa | u_i, \mathcal{M}_t, G_t)
\end{equation}
The retrieved skill $\kappa^*$ is then explicitly prepended to the generation context (visualized as the \texttt{<skill>} token), serving as a high-level directive for the subsequent response $r_i$.

\paragraph{Generative Skill Evolution.}
A key capability of the Skill Evolution Engine is its ability to update the skill tree after each session through a hierarchical, reflective process. This is enabled by fine-tuning the model on two specialized tasks designed to enforce structural consistency:
\begin{itemize}[leftmargin=*, align=left]
    \item \textbf{Targeted Skill Extraction:} rather than generic abstraction, the evolution is anchored in the session's therapeutic goal. The model first retrieves the most relevant Meta-Skill ($M$) aligned with the session goal. It then analyzes the high-reward trajectory $S^*$ to abstract a specific Atomic Skill ($\kappa_{atom}$) that instantiates this meta-strategy:
    \begin{equation}
        \kappa_{atom} \leftarrow \mathcal{E}_\theta(S^* \mid M, \text{Goal})
    \end{equation}
    This ensures that new skills are categorized under the correct pedagogical branch.
    
    \item \textbf{Adaptive Skill Management:} To maintain an efficient knowledge base, the model performs a context-aware update within the identified Meta-Skill branch. It retrieves the existing atomic skill ($\kappa_{ref}$) most semantically similar to $\kappa_{atom}$. The model then executes a decision policy $\pi_{mgmt}$ to either \textit{append} the new skill as a distinct node or \textit{merge} it with the existing one to refine the definition:
    \begin{equation}
        \mathcal{T}_{t+1} \leftarrow 
        \begin{cases} 
        \mathcal{T}_t \cup \{\kappa_{atom}\} & \text{if } \pi_{mgmt}(\kappa_{atom}, \kappa_{ref}) = \text{Append} \\
        \text{Merge}(\mathcal{T}_t, \kappa_{ref}, \kappa_{atom}) & \text{if } \pi_{mgmt}(\kappa_{atom}, \kappa_{ref}) = \text{Merge}
        \end{cases}
    \end{equation}
    This mechanism prevents redundancy while allowing the continuous refinement of existing techniques.
\end{itemize}
Analogous to the optimization strategy employed in the Memory module, we construct specialized SFT data to equip the agent with these evolutionary capabilities. Through this targeted fine-tuning, \texttt{PsychAgent} learns to autonomously curate its own professional knowledge base, realizing true open-ended learning.

\subsection{Reinforced Internalization Engine}
\label{sec:rie}

While the Skill Evolution Engine expands the agent's explicit knowledge base, the Reinforced Internalization Engine (Figure~\ref{fig:framework}c) focuses on transforming this external knowledge into endogenous intuition. To achieve this, we employ an iterative Rejection Fine-Tuning (RFT) mechanism~\citep{yuan2023scaling}, which mimics the human process of reflecting on and reinforcing successful counseling experiences while discarding ineffective ones.

\paragraph{Iterative Trajectory Rollout and Selection.}
Unlike standard reinforcement learning that explores at the token level, our engine performs exploration at the \textit{session level} to ensure coherent long-term strategies. As illustrated in the bottom panel of Figure~\ref{fig:framework}, the process unfolds iteratively for a sequence of sessions:
\begin{itemize}[leftmargin=*, align=left] 
    \item \textbf{Parallel Rollout:} At any session step $t$, given the history of the previous optimal sessions $\mathcal{H}^*_{t-1}$, the agent generates $N$ parallel candidate sessions $\{S_t^{(1)}, \dots, S_t^{(N)}\}$. Each candidate explores different combinations of retrieved skills and dialogue strategies.
    \item \textbf{Reward Evaluation:} We employ a comprehensive Reward Model to assess the quality of each completed session candidate, denoted as $S_t^{(k)}$. To capture the complexity of real-world counseling, the evaluation metric $\mathcal{R}$ is designed as a holistic integration of client- and counselor-centric dimensions using LLMs-as-a-Judge \citep{gu2024survey}, following \citep{pan2026psycheval}. The final reward score serves as a critical feedback signal, guiding the model toward strategies that are both clinically valid and psychologically supportive.
    \item \textbf{Greedy Selection:} We select the optimal session $S_t^*$ that yields the maximum reward: $S_t^* = \operatorname{argmax}_k \mathcal{R}(S_t^{(k)})$. Crucially, the memory state for the subsequent session $t+1$ is updated based \textit{only} on this optimal trajectory $S_t^*$. This ensures that the agent's lifelong learning path is constructed solely from its most successful interactions (marked with trophies in Figure~\ref{fig:framework}).
\end{itemize}

\paragraph{Internalization via Rejection Fine-Tuning.}
After constructing the optimal lifelong trajectory $\mathcal{T}^* = \{S_1^*, S_2^*, \dots, S_T^*\}$, the engine utilizes this data for parameter optimization. We treat the rejected sessions (those with lower rewards, marked with crosses) as negative samples implicitly. The model is fine-tuned to maximize the likelihood of the selected "golden" trajectory $\mathcal{T}^*$.
\begin{equation}
    \mathcal{L}_{RFT}(\theta) = - \sum_{t=1}^{T} \log P_\theta(S_t^* | \mathcal{H}^*_{t-1})
\end{equation}
Through this process, the explicit skills and planning logic used to generate $S_t^*$ are gradually "distilled" into the model's weights $\theta$. Over time, this allows \texttt{PsychAgent} to produce high-quality responses intuitively, reducing its dependence on heavy external retrieval.

\section{Experimental Results}
\subsection{Experimental Setups}

\paragraph{Datasets and Metrics.}
We build our experiments on \texttt{PsychEval}~\citep{pan2026psycheval}. Following its protocol, we first perform supervised fine-tuning (SFT) on the released multi-session counseling corpus, and then reuse its pool of 2,000+ client profiles for both rollout training and evaluation. For each therapeutic school, we sample 140 client profiles, using 120 for training and 20 for evaluation.
We adopt the evaluation protocol of \texttt{PsychEval}. Across all therapeutic schools, we report the shared counseling metrics, including \textsc{PANAS}, \textsc{RRO}, \textsc{SRS}, \textsc{Custom\_Dim}, \textsc{HTAIS}, \textsc{WAI}, and \textsc{Dialogue\_Planning}. In addition, each school is evaluated with two school-specific metrics: behavioral therapy with \textsc{MITI} and \textsc{STAI}, cognitive-behavioral therapy with \textsc{CTRS} and \textsc{BDI\_II}, postmodernist therapy with \textsc{EFT\_TFS} and \textsc{SFBT}, humanistic-existential therapy with \textsc{TES} and \textsc{CCT}, and psychodynamic therapy with \textsc{PSC} and \textsc{IPO}. Following \texttt{PsychEval}, we report both counselor-side and client-side results, further organized into therapy-shared and therapy-specific dimensions.

\paragraph{Baselines.}
We compare \texttt{PsychAgent} against two categories of baselines to disentangle the effects of general reasoning ability and domain-specific counseling specialization.
For \textbf{General-purpose LLMs}, we include several strong off-the-shelf LLMs: GPT-5.4~\citep{openai2026gpt54}, Gemini-3~\citep{google2025gemini3}, Qwen3-Max~\citep{yang2025qwen3}, and DeepSeek-V3.2~\citep{liu2025deepseek}. These baselines represent state-of-the-art general models and allow us to assess whether strong counseling performance can be achieved from general capability alone.
For \textbf{Psychology-Specific LLMs}, we include PsyLLM~\citep{lai2023psy}, PsyDTLLM~\citep{xie2025psydt}, CPsyCounX~\citep{zhang2024cpsycoun}, and TheraMind~\citep{hu2026theramindstrategicadaptiveagent}. PsyLLM incorporates explicit diagnostic and therapeutic reasoning, PsyDTLLM models counselor-specific style via digital twins, CPsyCounX is fine-tuned on reconstructed counseling dialogues, and TheraMind is designed for longitudinal counseling with cross-session planning and adaptation. 

\paragraph{Implementation Details.}
We finetune \texttt{PsychAgent} with Qwen3-32B and set the maximum context length to 32,768 tokens. 
All models are trained with DeepSpeed ZeRO-3 in \texttt{bf16} precision, using a per-device batch size of 1, gradient accumulation of 8, a learning rate of $1\times10^{-5}$, cosine decay, and 10\% warmup for 3 epochs. The rollout number $N=8$. 
History masking is enabled in all experiments unless otherwise specified. Training is conducted on two servers, each equipped with 8 NVIDIA H200 GPUs.


\begin{table}
\centering
\setlength{\tabcolsep}{6pt}
\renewcommand{\arraystretch}{1.0}
\caption{Main results in terms of counselor- and client-level metrics. $^\dag$ means the model is based on Qwen3-8B. Best results are in bold and second-best results are underlined.}
\label{tab:main}
\begin{tabular}{lcccc}
\toprule
\multirow{2}{*}{Model} 
& \multicolumn{2}{c}{Counselor-level} 
& \multicolumn{2}{c}{Client-level} \\
\cmidrule(lr){2-3}\cmidrule(lr){4-5}
& \makecell[c]{Shared}
& \makecell[c]{Specific}
& \makecell[c]{Shared}
& \makecell[c]{Specific} \\
\midrule
\multicolumn{5}{@{}l}{\textit{General-purpose LLMs}} \\
GPT-5.4       & 5.54 & 7.41 & 5.07 & 7.72 \\
Gemini-3      & 5.34 & 7.04 & 4.97 & 7.52 \\
Qwen3-Max     & 5.88 & 7.74 & 5.41 & 7.81 \\
DeepSeek-V3.2 & 5.54 & 7.12 & 5.06 & 7.70 \\
\midrule
\multicolumn{5}{@{}l}{\textit{Psychology-specific LLMs}} \\
PsyLLM        & 5.30 & 4.67 & 5.63 & 7.93 \\
PsyDTLLM      & 6.10 & 5.43 & 5.27 & 7.42 \\
CPsyCounX     & 4.21 & 2.51 & 4.73 & 7.27 \\
TheraMind     & 6.25 & 6.94 & 5.48 & 7.83 \\
\midrule
\rowcolor{black!4}
\texttt{PsychAgent}$^\dag$
& \best{7.35} & \second{7.78} & \best{5.94} & \second{8.19} \\
\rowcolor{black!4}
\texttt{PsychAgent}
& \second{7.32} & \best{7.91} & \second{5.92} & \best{8.24} \\
\bottomrule
\end{tabular}
\end{table}

\subsection{Main Results}
As shown in Table~\ref{tab:main}, \texttt{PsychAgent} achieves higher scores than both general-purpose and psychology-specific baselines across all four aggregated evaluation dimensions. Compared with the strongest general-purpose baseline, Qwen3-Max, \texttt{PsychAgent} improves counselor-level shared/specific scores by +1.44/+0.17 and client-level shared/specific scores by +0.51/+0.43. Relative to the longitudinal counseling agent TheraMind, the corresponding gains are +1.07, +0.97, +0.44, and +0.41 on the four dimensions. These results suggest that strong general reasoning or longitudinal planning alone may not fully account for high counseling performance on this benchmark.The gains remain substantial against the longitudinal counseling agent TheraMind, with improvements of +1.07, +0.97, +0.44, and +0.41 on the four dimensions. 
These results suggest that the observed gains are not explained solely by domain adaptation or cross-session planning. Notably, even \texttt{PsychAgent}$^\dag$ (8B) achieves the best or second-best result on every dimension, which suggests that the improvements are associated with the proposed lifelong learning framework in addition to model scale.
Figure~\ref{fig:radar} in the appendix further suggests that the gains are broad-based rather than concentrated in only one or two metrics. The largest improvements appear on counselor-side planning and intervention measures, including \textsc{Dialogue\_Planning} (9.41 vs. 6.52) and \textsc{CTRS} (9.41 vs. 7.96), consistent with stronger session structuring and more theory-aligned interventions under this benchmark. Meanwhile, steady gains are also observed on client-side outcomes such as \textsc{CCT} (8.13 vs. 7.04) and \textsc{SFBT} (6.71 vs. 5.57), suggesting that the benefits of the framework are reflected not only in counselor behavior but also in client-side outcome measures. Overall, these results are consistent with effective counseling in this benchmark benefiting from both longitudinal context handling and continued skill evolution/internalization.
\begin{wraptable}{r}{0.5\textwidth}
\vspace{-3mm}
\centering
\caption{Results of human evaluation.}
\label{tab:human-eval-desc}
\vspace{2mm}
\setlength{\tabcolsep}{1.5pt}
\begin{tabular}{lcccc}
\toprule
\textbf{Model} & \textbf{Human A} & \textbf{Human B} & \textbf{LLM} & \textbf{Rank} \\
\midrule
PsychAgent & \textbf{4.295} & \textbf{4.370} & \textbf{4.303} & \textbf{1} \\
Qwen3-Max & 3.943 & 4.024 & 4.039 & 2 \\
TheraMind & 3.743 & 3.649 & 3.573 & 3 \\
\bottomrule
\end{tabular}
\end{wraptable}
\subsection{Human Evaluation}
\label{sec:human-eval}

To complement automatic evaluation, we conduct human expert evaluation on multi-session counseling dialogues generated by \texttt{PsychAgent}, Qwen3-Max, and TheraMind. Each dialogue is independently rated by two human annotators and one LLM rater (Gemini-3) on four dimensions \textit{Ethics}, \textit{Interaction}, \textit{Intervention}, and \textit{Perception} using a 1–5 ordinal scale, and we report Quadratic Weighted Kappa (QWK)~\citep{cohen1968weighted} to measure agreement. Across 522 matched dialogues, \texttt{PsychAgent} is ranked first by both human raters and the LLM rater, suggesting that its relative advantage is not limited to automatic metrics. Meanwhile, annotator agreement is moderate to strong: human–human agreement reaches 0.675 QWK, while the LLM rater attains higher agreement with the two human raters (0.770 and 0.877), suggesting that it may serve as a useful auxiliary evaluator in this setting.
Due to space constraints, the annotation guidelines, annotation details, and detailed analyses of the evaluations are provided in Section~\ref{sec:appendix-human-eval} of the appendix.

\begin{wraptable}{r}{0.5\textwidth}
\vspace{-5mm}
\centering
\caption{The results of ablation studies.}
\label{tab:ablation}
\vspace{2mm}
\setlength{\tabcolsep}{2mm}
\begin{tabular}{lcccc}
\hlineB{4}
  & \multicolumn{2}{c}{Counselor-level} & \multicolumn{2}{c}{Client-level}  \\
                         & Shared  & Specific  & Shared & Specific \\ \hline
 \texttt{PsychAgent}  & \textbf{7.32} & \textbf{7.91} & \textbf{5.92} & \textbf{8.24} \\ 
w/o MAPE     & 7.07 & 7.66 & 5.86 & 8.03 \\
w/o SEE      & 7.02 & 7.52 & 5.61 & 7.90 \\
w/o RIE      & 7.05 & 7.67 & 5.67 & 7.89 \\
\hlineB{4}
\end{tabular}
\end{wraptable}

\subsection{Ablation Studies}
Table~\ref{tab:ablation} suggests that all three components contribute to the final performance of \texttt{PsychAgent}. Removing any module consistently lowers both counselor-level and client-level results, which is consistent with the overall gains arising from their combination. Among them, removing the Skill Evolution Engine (SEE) leads to the largest drop in this ablation, suggesting that skill evolution may be the most influential component under the current setting. Removing the Reinforced Internalization Engine (RIE) also causes noticeable performance loss, indicating that retrieved skills benefit from further internalization. Removing the Memory-Augmented Planning Engine (MAPE) results in a smaller but still consistent decline, highlighting the contribution of longitudinal memory and session-level planning. Overall, the results suggest that strong psychological counseling requires not only memory, but also skill acquisition and internalization from experience.


\subsection{Further Analysis}
\begin{wrapfigure}{r}{0.6\textwidth}
\vspace{-5mm}
    \centering
    \includegraphics[width=0.6\textwidth]{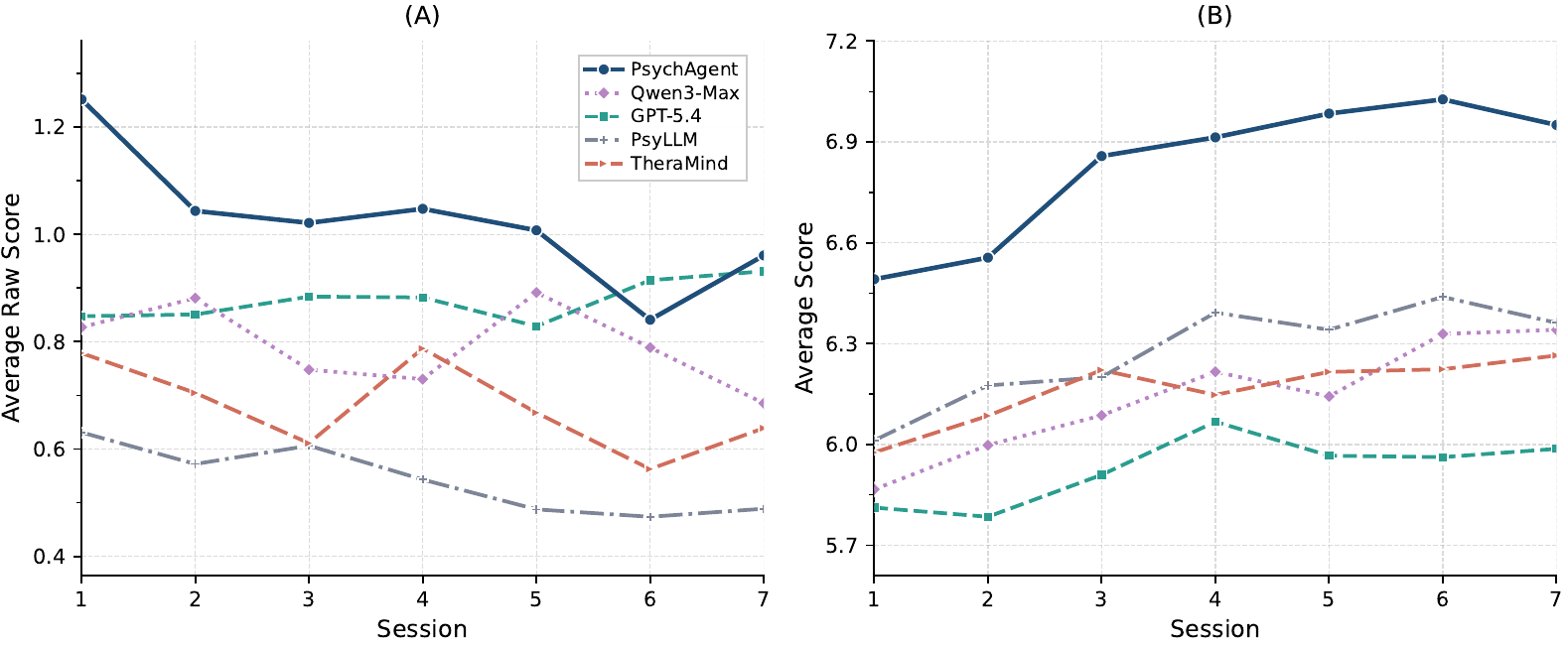}
    \vspace{-6mm}
    \caption{Clients' Emotional Trajectories.}
    \label{fig:trend}
\vspace{-3mm}
\end{wrapfigure}


\paragraph{Analysis of Clients' Emotional Trajectories.}
Figure~\ref{fig:trend} compares the session-wise client trajectories across models. In Panel (A), where lower values are better, \texttt{PsychAgent} shows a relatively pronounced downward trend, suggesting a larger reduction in negative client states over time. In Panel (B), where higher values are better, it remains above the compared baselines across much of the trajectory and improves steadily across sessions. Relative to the flatter or less stable trajectories of other models, these patterns are consistent with more stable multi-session performance, although they should be interpreted as benchmark trends rather than direct clinical evidence.

\begin{wraptable}{r}{0.5\textwidth}
\vspace{-5mm}
\centering
\caption{Influence of Memory Mechanisms.}
\label{tab:memory}
\vspace{2mm}
\setlength{\tabcolsep}{2.0mm}
\begin{tabular}{lcccc}
\hlineB{4}
  & \multicolumn{2}{c}{Counselor-level} & \multicolumn{2}{c}{Client-level}  \\
                         & Shared  & Specific  & Shared & Specific \\ \hline
DeepSeek-V3.2                  & 5.54 & 7.12 & 5.06 & 7.70 \\ 
+ Vanilla RAG                  & 5.45 & 6.79 & 4.92 & 7.67 \\
+ Graph RAG                    & 5.45 & 6.88 & 4.92 & 7.73 \\
+ Mem0                         & 5.45 & 6.94 & 4.87 & 7.68 \\
+ MemoryBank                   & 5.49 & 6.92 & 4.90 & 7.70 \\
                               \hline
\texttt{PsychAgent}            & 7.32 & 7.91 & 5.92 & 8.24 \\
\hlineB{4}
\end{tabular}
\end{wraptable}

\paragraph{Influence of Memory Mechanisms}
Table~\ref{tab:memory} suggests that simply adding existing memory architectures to a strong backbone such as DeepSeek-V3.2 yields limited benefit for this benchmark, and in some cases slightly lowers performance. This pattern indicates that generic memory methods, including Vanilla RAG~\citep{lewis2020retrieval}, Graph RAG~\citep{edge2024local}, Mem0~\citep{chhikara2025mem0}, and MemoryBank~\citep{zhong2024memorybank}, may be insufficient on their own for this setting. One possible explanation is that counseling requires more than recalling past facts: the model may also need to track evolving mental states, therapeutic progress, and the rationale behind prior interventions. In contrast, \texttt{PsychAgent} achieves higher scores across all metrics, which is consistent with the value of combining memory with skill learning and internalization.


\paragraph{Case Studies}
To better understand why \texttt{PsychAgent} improves counseling performance, we examine representative cases of skill extraction and post-internalization behavior (Appendix \ref{sec:case_study}). Three observations are particularly notable. First, the model often distills practice-grounded skills from counseling trajectories rather than merely restating entries in the initial skill library, which is consistent with a form of self-evolution from practice (Section \ref{sec:case_studies_skill_extraction}). Second, after internalization, the model often combines existing micro-skills into \emph{new functional skills} rather than only reusing predefined ones, suggesting a degree of skill-level compositional reorganization beyond the original inventory (Section \ref{sec:operational_new_functional_skills}). Third, some internalized skill packages recur across therapeutic orientations, suggesting partial cross-context reuse of intervention patterns rather than strict therapy-specific reuse only (Section \ref{sec:cross_therapy_reuse}). Taken together, these findings indicate broader functional reuse, though not therapy-agnostic counseling.

\section{Conclusions and Further Work}
We introduced \texttt{PsychAgent}, an experience-driven lifelong learning framework for AI psychological counseling that integrates memory-augmented planning, skill evolution, and reinforced internalization. By moving beyond static fine-tuning, \texttt{PsychAgent} enables the counselor agent to improve through accumulated experience and deliver more consistent, professional, and effective multi-session support. Experiments across multiple therapeutic schools show that our framework consistently outperforms both strong general-purpose LLMs and psychology-specific baselines. Future work will focus on evaluating the framework in more realistic counseling settings and strengthening safety and privacy protections for real-world deployment.



\bibliographystyle{unsrt}  
\bibliography{references}  


\appendix

\begin{figure}[h]
    \centering
    \includegraphics[width=0.99\textwidth]{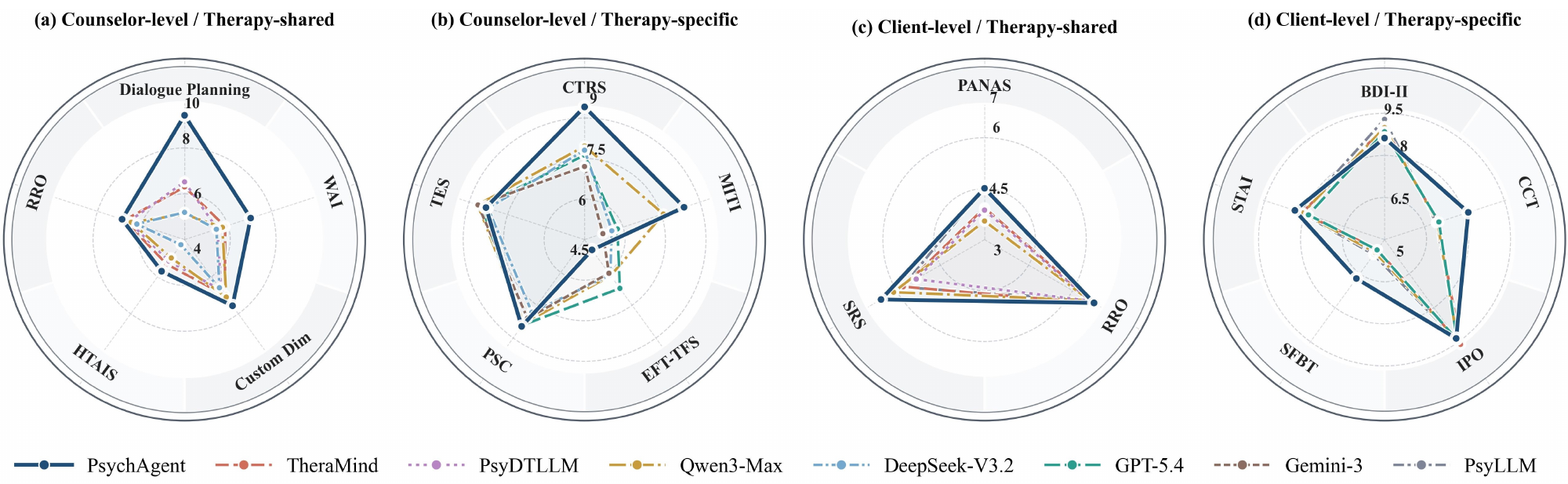}
    \caption{Detailed results across Counselor- and Client-Level therapy metrics}
    \label{fig:radar}
\end{figure}

\section{Detailed Results Across Counselor- and Client-Level Therapy Metrics}
Figure~\ref{fig:radar} provides a fine-grained comparison across counselor- and client-level therapy metrics, covering both therapy-shared and therapy-specific dimensions. Across all four panels, \texttt{PsychAgent} exhibits a consistently strong performance profile rather than isolated improvements on only a few metrics. The largest gains appear on counselor-side planning and intervention dimensions, especially \textsc{Dialogue\_Planning} and \textsc{CTRS}, suggesting stronger ability to structure therapeutic progress and deliver theory-consistent interventions. Meanwhile, the improvements also extend to client-side outcome measures such as \textsc{CCT} and \textsc{SFBT}, indicating that the benefits of the framework are reflected not only in counselor behavior but also in downstream client-level therapeutic progress. Overall, the radar plots support the conclusion that the proposed lifelong learning framework yields broad and coherent improvements across therapy dimensions, with especially clear advantages on metrics that depend on accumulated therapeutic skills and their effective deployment in ongoing dialogue.

\section{Human Evaluation}
\label{sec:appendix-human-eval}
\subsection{Raters and Annotation Protocol}
We conducted human evaluation on multi-session counseling dialogues generated by PsychAgent, Qwen3-Max, and TheraMind. The two human raters were graduate students with psychology-related training, and the third evaluator was LLM Rater (Gemini-3). All three raters evaluated the \emph{counselor's performance}, rather than the client's state, on four dimensions: \textit{Ethics}, \textit{Interaction}, \textit{Intervention}, and \textit{Perception}. All scores were assigned on a \mbox{1--5} ordinal scale. Because these labels are ordered categorical variables, we report Quadratic Weighted Kappa (QWK) as the primary agreement metric; QWK is more appropriate than unweighted agreement measures because it accounts for the ordinal distance between adjacent scores.
\begin{casebox}{Human Evaluation Rubric and Annotation Principles}
\small
\begin{tabularx}{\linewidth}{@{}p{0.24\linewidth}X@{}}
\textbf{Evaluation target} &
Rate the \emph{counselor's performance}, rather than the client's state, on four dimensions: \textit{Ethics}, \textit{Interaction}, \textit{Intervention}, and \textit{Perception}. \\

\textbf{General principles} &
Ratings must rely only on information explicitly stated in the dialogue; unsupported inference is disallowed. A score of 3 serves as the default when evidence is limited, performance is unremarkable, or the relevant ability is not clearly manifested. Scores of 4--5 require clear supporting evidence, and a score of 5 is used sparingly. Obvious risks, boundary violations, misreadings, and unsafe responses lower the score. Opportunity-sensitive abilities are not penalized when the relevant scenario is absent. \\

\textbf{Ethics} &
Whether the counselor responds safely, respects boundaries, avoids misleading diagnosis or labeling, and does not provide clearly unsafe or inappropriate guidance. \\

\textbf{Interaction} &
Whether the counselor maintains a coherent, natural, and responsive conversation, avoids obvious repetition or non sequiturs, and moves the dialogue forward at an appropriate pace and tone. \\

\textbf{Intervention} &
Whether the counselor provides help that is appropriate to the client's current problem and state, and whether the guidance is clear, feasible, and practically usable. \\

\textbf{Perception} &
Whether the counselor accurately understands and responds to the client's emotions and psychological state, conveys empathy, and helps the client better understand themselves or see possible resources, hope, or direction. \\

\textbf{Scoring rule} &
1 = clearly poor; 2 = weak; 3 = acceptable, unremarkable, or insufficient evidence; 4 = good with clear strengths and no major problems; 5 = very good, used conservatively and only when multiple clear strengths are supported by explicit evidence. \\

\textbf{Evidence recording} &
For each dimension, raters provide one overall 1--5 score, one key positive evidence span, and, when applicable, one negative evidence span in a structured format. \\
\end{tabularx}

\vspace{0.4em}
\textbf{Why this matters.}
This rubric enforces evidence-grounded and conservative scoring, reduces annotation ambiguity, and makes human--LLM comparisons more interpretable and reproducible.
\end{casebox}

The matched evaluation set contains 522 fully rated dialogues from 25 cases and 188 case-session units. This yields 6,264 dimension-level observations and 2,088 paired ratings for each rater pair. One dialogue was excluded because all three rating files failed for the same item, leaving a nearly complete matched sample. The sample is balanced across counselor models (173 PsychAgent dialogues, 175 Qwen3-Max dialogues, and 174 TheraMind dialogues), which supports comparison at the overall, by-dimension, by-model, and model-by-dimension levels.

\subsection{Inter-Rater Agreement}

Table~\ref{tab:human-eval-desc} shows that all three raters concentrate heavily on high scores, with \mbox{4--5} accounting for 78.8\%--81.7\% of all dimension-level ratings. This indicates a generally favorable score distribution and also suggests some degree of ceiling effect, especially for \textit{Perception}. Human Rater B is slightly more lenient than Human Rater A overall, whereas LLM Rater (Gemini-3) exhibits the largest spread and uses low scores somewhat more frequently. Nevertheless, the relative ranking of the three counselor models is remarkably stable across raters: PsychAgent performs best, Qwen3-Max is consistently second, and TheraMind performs worst.

\begin{table*}[t]
\centering
\small
\setlength{\tabcolsep}{3pt}
\begin{tabular}{llcccc}
\toprule
\textbf{Level} & \textbf{Group} & \textbf{$n$} & \textbf{Human A vs Human B} & \textbf{LLM vs Human A} & \textbf{LLM vs Human B} \\
\midrule
Overall & All ratings & 2088 & 0.675 & 0.770 & 0.877 \\
\midrule
\multirow{4}{*}{By dimension}
& Ethics & 522 & 0.606 & 0.756 & 0.823 \\
& Interaction & 522 & 0.576 & 0.691 & 0.841 \\
& Intervention & 522 & 0.747 & 0.812 & 0.907 \\
& Perception & 522 & 0.546 & 0.660 & 0.843 \\
\midrule
\multirow{3}{*}{By model}
& PsychAgent & 692 & 0.626 & 0.752 & 0.865 \\
& Qwen3-Max & 700 & 0.698 & 0.813 & 0.854 \\
& TheraMind & 696 & 0.528 & 0.649 & 0.821 \\
\bottomrule
\end{tabular}
\caption{Quadratic Weighted Kappa (QWK) across raters.}
\label{tab:human-eval-qwk}
\end{table*}

Agreement results are reported in Table~\ref{tab:human-eval-qwk}. At the overall level, Human Rater A vs Human Rater B achieves a QWK of 0.675, indicating moderately strong agreement. In contrast, LLM Rater (Gemini-3) attains higher agreement with both human raters: 0.770 against Human Rater A and 0.877 against Human Rater B. Exact agreement follows the same trend (74.2\% for Human A vs Human B, 78.7\% for LLM vs Human A, and 90.3\% for LLM vs Human B), and nearly all disagreements are within one rating point. These results suggest that the LLM rater is not behaving randomly; rather, it tracks a substantial portion of the human scoring structure, especially the rating style of Human Rater B.

The agreement pattern is also dimension-dependent. Across all three pairings, \textit{Intervention} is the most reliable dimension, reaching QWK values of 0.747, 0.812, and 0.907. By contrast, \textit{Perception} shows the weakest human-human agreement (0.546), and \textit{Interaction} is also comparatively variable. This is consistent with the descriptive distributions: \textit{Perception} is highly concentrated in the upper score range, which likely compresses effective variance and makes Kappa estimates more sensitive to small shifts in rating style.

\subsection{Systematic Bias of the LLM Rater}

\begin{table*}[t]
\centering
\small
\setlength{\tabcolsep}{7pt}
\begin{tabular}{llccc}
\toprule
\textbf{Level} & \textbf{Group} & \textbf{$n$} & \textbf{LLM $-$ Human A} & \textbf{LLM $-$ Human B} \\
\midrule
Overall & All ratings & 2088 & $-0.022$ & $-0.043$ \\
\midrule
\multirow{4}{*}{By dimension}
& Ethics & 522 & $-0.031$ & $-0.102$ \\
& Interaction & 522 & \phantom{$-$}0.025 & $-0.048$ \\
& Intervention & 522 & $-0.042$ & $-0.038$ \\
& Perception & 522 & $-0.038$ & \phantom{$-$}0.017 \\
\midrule
\multirow{3}{*}{By model}
& PsychAgent & 692 & \phantom{$-$}0.009 & $-0.066$ \\
& Qwen3-Max & 700 & \phantom{$-$}0.096 & \phantom{$-$}0.014 \\
& TheraMind & 696 & $-0.170$ & $-0.076$ \\
\bottomrule
\end{tabular}
\caption{Systematic mean-score bias of LLM Rater (Gemini-3) relative to each human rater. Positive values indicate that the LLM assigns higher scores on average.}
\label{tab:human-eval-bias}
\end{table*}

Table~\ref{tab:human-eval-bias} shows that LLM Rater (Gemini-3) does not exhibit a simple global tendency to inflate scores. Relative to Human Rater A, the overall mean bias is nearly zero (\textbf{$-0.022$}), but this average conceals substantial model-specific structure: the LLM is slightly more favorable on PsychAgent and Qwen3-Max, yet clearly harsher on TheraMind ($-0.170$). Relative to Human Rater B, the LLM is more consistently conservative overall ($-0.043$), with its largest dimension-level deviation on \textit{Ethics} ($-0.102$). Thus, the LLM appears to behave like a relatively stable additional evaluator whose judgments are closer to Human Rater B than to Human Rater A, rather than like a neutral proxy for an underlying gold standard.

Overall, these results suggest that, under a shared and evidence-grounded rubric, LLM Rater (Gemini-3) provides \emph{substantial practical reliability} as an auxiliary evaluator in this setting: it recovers the same model ranking as human raters, achieves high QWK against both humans, and disagrees mostly by only one rating point. However, its alignment is not uniform across dimensions or counselor models. In particular, \textit{Ethics}, \textit{Perception}, and lower-performing models such as TheraMind remain more sensitive to rating-style differences. We therefore view the LLM rater as a scalable complement to human evaluation rather than a drop-in replacement for multi-rater human judgment.

\section{Case Studies}
\label{sec:case_study}

\subsection{Extracting New Skills from Conversations}
\label{sec:case_studies_skill_extraction}

To better understand why \texttt{PsychAgent} improves counseling performance, we examine whether its extracted skills merely restate the initial skill library or instead capture \emph{new practice-grounded skill units} from real counseling dialogue. Here, we define a practice-grounded skill unit as an extracted skill that introduces \emph{non-trivial practical increments} beyond the closest existing micro-skills, rather than simply rephrasing or elaborating pre-existing entries. Concretely, the comparison baseline is the \texttt{v3} library within the same therapeutic orientation and stage, where micro-skills are already fine-grained and typically include \texttt{when\_to\_use} and \texttt{trigger} fields. As a result, a skill is counted as novel only when it adds practical structure that is not already captured by the nearest existing micro-skills.

Under this criterion, the answer is clearly positive. Across 715 dialogue records, the extraction pipeline produces 9,701 atomic skills, among which 5,923 qualify as practice-grounded skill units (\texttt{B}+\texttt{C}). This pattern holds across all five orientations: the proportion of practice-grounded skill units reaches 73.29\% in BT, 69.77\% in CBT, 56.0\% in HET, 51.61\% in PDT, and 63.49\% in PMT. These results show that the dominant outcome of extraction is not restatement, but the distillation of practically enriched skill units relative to the closest existing micro-skills.

The practical increments are also highly consistent across orientations. The most frequent additions are clearer operational sequencing, minimum viable actions, quantitative thresholds, structured templates, monitoring or recording loops, and pause or rollback rules. In other words, the model does not mainly invent completely new therapy techniques; instead, it absorbs interactional experience from counseling trajectories and re-expresses it as more executable local skill units. At the same time, our stricter global audit finds no sufficiently supported candidates for \emph{entirely novel therapy techniques} relative to the initial library. Even the strongest cases are better understood as stable repackagings of existing techniques with added thresholds, templates, monitoring rules, or minimum-action designs.

Representative cases from all five orientations further illustrate this pattern.

\begin{casebox}{BT: Micro-Contracts Through Thresholds and Flexible Logging}
\small
\begin{tabularx}{\linewidth}{@{}p{0.22\linewidth}X@{}}
\textbf{Closest library skills} &
Formalize and sign a behavioral contract (\texttt{661}); specify monitoring arrangements in the contract (\texttt{662}); implement reinforcement and punishment contingencies as specified (\texttt{663}). \\
\textbf{Representative extracted skill} &
\emph{Rule-based micro-contract with thresholds and flexible logging}. \\
\textbf{Practical increment} &
The extracted unit turns a generic contract into a concrete execution package: an explicit red-light threshold, a fallback sequence, a fixed minimum practice duration, and a logging option adapted to the client's preferred emoji-based format. \\
\end{tabularx}

\vspace{0.4em}
\textbf{Dialogue evidence.}
\begin{quote}\small
\textbf{Client:} This week I tried audio recording, but halfway through it felt awkward, like talking to myself, so I switched to sending cat emojis to my writing buddy. After sending them, the pressure felt lower. When you suggested ``just write the simplest sentence,'' yesterday I only wrote ``The protagonist stared at the balcony for five minutes,'' and somehow ended up writing half a page. \\
\textbf{Counselor:} ``Just write the simplest sentence'' often creates unexpected fluency. Let us write the rule clearly: if tension is $\geq 7/10$, first bring it down to $\leq 3/10$, and only then write for 10 minutes; for logging, use ``practice/down/write'' or the cat emojis you prefer. Would that work? \\
\textbf{Client:} I think that could work. Honestly, sometimes seeing ``practice/down/write'' suddenly irritates me, as if I am boxed into another task. But recording with cat emojis might feel easier.
\end{quote}

\textbf{Interpretation.}
This case goes beyond recovering a generic BT label such as behavioral contracting. The extracted skill adds an explicit threshold, a rollback rule, a minimum viable practice target, and a client-preferred logging format. The non-trivial increment lies in turning a broad contract into a locally executable self-management unit grounded in the client's actual resistance and workaround.
\end{casebox}

\begin{casebox}{CBT: One-Page Process Cards with Low-Threshold Activation}
\small
\begin{tabularx}{\linewidth}{@{}p{0.22\linewidth}X@{}}
\textbf{Closest library skills} &
Attribute progress to the client to strengthen self-efficacy (\texttt{1247}); review what was learned and apply it in daily life (\texttt{1433}); emphasize that tools can transfer across situations and encourage continued use (\texttt{1248}). \\
\textbf{Representative extracted skill} &
\emph{Design a one-page process card with a low activation threshold and a personalized starter cue}. \\
\textbf{Practical increment} &
The extracted unit compresses multiple CBT elements into a reusable local workflow: step order, an optional low-burden record, a user-specific cue phrase, and concrete in-situation calibration actions. \\
\end{tabularx}

\vspace{0.4em}
\textbf{Dialogue evidence.}
\begin{quote}\small
\textbf{Counselor:} Let us organize the tools into a ``one-page process card.'' The skeleton is: pause when tension rises $\rightarrow$ two slow breaths $\rightarrow$ glance at one sentence on the card $\rightarrow$ calibrate one body part $\rightarrow$ return to the beat $\rightarrow$ do a before/after rating. In class, would you be willing to track ``beats aligned'' and ``toe calibrations'' as easy indicators? What sentence should we add that best fits you? \\
\textbf{Client:} This week I tried tracking beat alignment a few times, and it worked better than just feeling anxious. But sometimes even glancing at the card makes me think, ``Is recording numbers too deliberate?'' What about adding, ``Just get one beat right first''? I often panic on the very first beat. \\
\textbf{Counselor:} That concern is realistic. Let us keep the record ``optional + observational'': either write one fact or just record the beat count, with no performance target. ``Just get one beat right first'' fits your sticking point well, so let us put it on the front of the card.
\end{quote}

\textbf{Interpretation.}
This case is not merely a more detailed restatement of tool use or review. The extracted skill introduces a stable mini-workflow with sequencing, low-threshold activation, and a cue phrase stated in the client's own language. What is distilled from practice is not just ``use CBT tools,'' but how to make those tools deployable when the client hesitates at the point of initiation.
\end{casebox}

\begin{casebox}{HET: SMART Self-Care Goals with Sequenced Choice and Logging}
\small
\begin{tabularx}{\linewidth}{@{}p{0.22\linewidth}X@{}}
\textbf{Closest library skills} &
Guide an initial choice toward action (\texttt{863}); help the self become experienced as choosing (\texttt{861}); organize wishes into will (\texttt{858}). \\
\textbf{Representative extracted skill} &
\emph{Convert self-care intentions into SMART goals and concrete logging routines}. \\
\textbf{Practical increment} &
The extracted unit turns broad existential choice into a sequenced action package: validate the wish, bridge it to one prioritized domain, specify a SMART target, and attach a small logging procedure for follow-through. \\
\end{tabularx}

\vspace{0.4em}
\textbf{Dialogue evidence.}
\begin{quote}\small
\textbf{Client:} For sleep I do want to try getting in bed before 11, but before sleep I always end up scrolling on my phone. A brisk walk at lunch might be okay for exercise, but for diet I am hesitant; changing that all at once feels stressful. \\
\textbf{Client:} Actually I often scroll on my phone very late and feel exhausted the next day. This week I want to try putting the phone down before 11 and listening to light music before sleep. I probably cannot do it every day right away; maybe start with three days? \\
\textbf{Counselor:} We could define the sleep goal as ``Choose three days this week to put the phone down before 11 and listen to 10 minutes of light music before sleep.'' Does that sound right? At the same time, would you be willing to set one small goal and one recording method for exercise and diet as well?
\end{quote}

\textbf{Interpretation.}
This case shows that the extraction goes beyond a generic HET emphasis on choice or commitment. The resulting skill unit encodes how abstract self-care wishes are translated into staged, measurable, and recordable action. The practical increment lies in making existential choice executable through prioritization, SMART specification, and lightweight monitoring.
\end{casebox}

\begin{casebox}{PDT: Minimal Viable Goals with Safety Boundaries}
\small
\begin{tabularx}{\linewidth}{@{}p{0.22\linewidth}X@{}}
\textbf{Closest library skills} &
Collaboratively formulate goals from the client's concerns (\texttt{180}); prioritize safety and urgent concerns while setting layered goals (\texttt{178}); jointly formulate short- and long-term counseling goals (\texttt{274}). \\
\textbf{Representative extracted skill} &
\emph{Collaboratively set a minimal viable target with simplified recording and a pause boundary}. \\
\textbf{Practical increment} &
The extracted unit refines broad collaborative goal-setting into a concrete local package: a minimum daily target, a reduced record format, a timing choice, and an explicit permission to pause. \\
\end{tabularx}

\vspace{0.4em}
\textbf{Dialogue evidence.}
\begin{quote}\small
\textbf{Client:} Doing this exercise after getting up in the morning might work best, because that is when my mind is still clear. But honestly I worry I will not keep it up. I already skipped the breathing practice for two days last time, and writing these records sometimes feels like reminding myself that I am anxious again. \\
\textbf{Counselor:} Let us cut the task to the minimum: this week, just guarantee one round of diaphragmatic breathing plus brief relaxation each day. For the ABC record, you can write only one sentence of the original thought. Which feels more reachable: one full ABC each day, or starting with just ``one original sentence''? \\
\textbf{Client:} Starting with one sentence would probably be easier to stick to. When I tried ABC last week, I got especially irritated at the B step and had no idea what to write. But the breathing really did help.
\end{quote}

\textbf{Interpretation.}
This case is not simply another instance of collaborative goal-setting. The extracted skill captures a practice-grounded shift from full compliance demands to a minimum viable target, a low-burden record, and an explicit safety boundary around overload. The novelty lies in operationalizing how to preserve adherence when reflective work itself becomes aversive.
\end{casebox}

\begin{casebox}{PMT: Strength Cards for Pre-Session Force-Field Preparation}
\small
\begin{tabularx}{\linewidth}{@{}p{0.22\linewidth}X@{}}
\textbf{Closest library skills} &
Reframe obstacles and encourage exception behaviors (\texttt{245}); use exception questions to strengthen positive forces (\texttt{244}); collaboratively map a goal force field (\texttt{242}). \\
\textbf{Representative extracted skill} &
\emph{Assign ``strength cards'' with a minimum quota and a fixed note-taking template to prepare force-field analysis}. \\
\textbf{Practical increment} &
The extracted unit converts broad exception/resource work into a concrete between-session preparation protocol: a minimum number of entries, required evidence categories, and a reusable recording template. \\
\end{tabularx}

\vspace{0.4em}
\textbf{Dialogue evidence.}
\begin{quote}\small
\textbf{Client:} Sure, but recording feels a bit troublesome. I only used that five-item template for two days last week and then forgot. But the interviewer really did say my last case analysis was good---does that count as external feedback? \\
\textbf{Client:} I think this arrangement is okay. Yesterday I sent out two resumes using an 80\% standard, and afterward I really did go make tea. What exactly goes on the ``strength card''? I am not quite sure yet. \\
\textbf{Counselor:} Let us make the ``strength card'' minimal: collect three this week---at least one piece of external feedback, one successful experience, and one value/preference statement. Use your phone notes and record them as ``situation---what you did---other people's or bodily response---what is reusable.'' Does that feel workable?
\end{quote}

\textbf{Interpretation.}
This case shows that extraction does more than recover a generic PMT idea such as identifying strengths or exceptions. The resulting unit adds a quota, evidence categories, and a fixed template, thereby transforming a broad resource-focused intervention into a structured pre-session data-gathering routine that can reliably feed later force-field analysis.
\end{casebox}

Taken together, these cases clarify what it means for skill extraction to produce new practice-grounded skill units. The resulting skills are not merely longer descriptions of existing entries; instead, they encode interaction-derived thresholds, monitoring rules, minimum viable actions, sequencing constraints, and reusable templates that are not fully specified by the closest existing micro-skills. This is why the strongest supported conclusion is not that the model invents entirely new therapy techniques, but that it reliably distills practically enriched skill units from counseling trajectories.

\begin{table}[t]
\centering
\caption{Overall structural comparison between the SFT baseline and internalization data. The main shift is not a larger share of multi-skill turns, but more stable functional packaging and clearer family-level reuse across contexts.}
\label{tab:internalization_overview}
\small
\setlength{\tabcolsep}{4pt}
\begin{tabular}{lcc}
\toprule
\textbf{Metric} & \textbf{SFT} & \textbf{Internalization} \\
\midrule
Dialogue / session count & 6701 & 729 \\
Counselor turns & 154{,}966 & 20{,}869 \\
Turns with \texttt{<skill>} tags & 154{,}959 & 20{,}865 \\
Unique matched multi-skill bundles & 19{,}296 & 2{,}396 \\
Isolated-skill turn ratio & 50.8\% & 65.8\% \\
Multi-skill turn ratio & 49.2\% & 34.2\% \\
Cross-therapy bundle turn ratio & 0.0\% & 0.6\% \\
\bottomrule
\end{tabular}
\end{table}

\subsection{Operational New Functional Skills}
\label{sec:operational_new_functional_skills}

To better understand why internalization improves counseling performance, we compare post-internalization behavior against both the initialized skill library and the SFT baseline. Two observations are central. First, internalization does not simply increase the number of multi-skill turns. Instead, it reorganizes existing micro-skills into more stable and functionally coherent turn-level packages. Second, some of these internalized packages are reused across therapies, suggesting that the model acquires transferable counseling principles rather than only therapy-specific templates. Together, these patterns indicate that internalization pushes the model toward a more general and versatile counselor.

The clearest evidence for internalization is not a mechanical increase in multi-skill turns, but the emergence of recurrent higher-order packages with clearer local intervention functions. In other words, the model does not merely place multiple skill labels in the same turn; rather, it increasingly organizes several existing micro-skills into compact intervention units that behave like \emph{new functional skills} at the operational level.

Importantly, these packages should not be interpreted as newly invented canonical techniques. A more conservative interpretation is that internalization compresses multiple known micro-skills into recurrent, stable, and functionally closed turn-level units. This is why the strongest evidence for compositional generalization comes from the quality and stability of these packages, rather than from raw bundle counts alone.

\begin{table}[t]
\centering
\caption{Representative candidate functional skills after internalization. These packages are best interpreted as operationally new functional units built from known micro-skills, rather than as entirely new canonical techniques.}
\label{tab:operational_functional_skills}
\small
\setlength{\tabcolsep}{4pt}
\begin{tabular}{l l c c c}
\toprule
\textbf{Functional skill} & \textbf{Therapy} & \textbf{Internal} & \textbf{SFT} & \textbf{Status} \\
\midrule
Goal-to-Record Bridge & CBT & 30 & 58 & stable \\
Rated Reappraisal Loop & CBT & 22 & 104 & stable \\
Dose-Specified Practice Scaffold & BT & 18 & 72 & stable \\
Validation-Wrapped Pacing Regulation & PDT & 6 & 0 & stable \\
Alternative-Story Amplification & PMT & 241 & 1338 & stable \\
Resource-to-Action Scaffold & HET & 54 & 166 & stable \\
\bottomrule
\end{tabular}
\end{table}

\begin{casebox}{CBT: Goal-to-Record Bridge}
\small
\begin{tabularx}{\linewidth}{@{}p{0.22\linewidth}X@{}}
\textbf{Constituent micro-skills} &
Negotiate and clarify near-term goals; assign logging homework to support later evaluation. \\
\textbf{Representative functional skill} &
\texttt{Goal-to-Record Bridge}. \\
\textbf{Frequency} &
Internalization: 30; SFT: 58. \\
\end{tabularx}

\vspace{0.4em}
\textbf{Dialogue evidence.}
\begin{quote}\small
\textbf{Client:} What I most want to address first is that awkward feeling. Every time the topic turns to these things, I feel intensely uncomfortable, almost as if I were being stripped bare. If I could care a little less about other people's judgment first, maybe I could gradually try to join such conversations. \\
\textbf{Counselor:} Let us make the goal more concrete: reduce the awkwardness from very strong to moderate or lighter, and use the number of times you eat alone as a trackable indicator. At the same time, I suggest keeping a trigger diary each day, recording the situation, the automatic thought, the emotion intensity, and what you did. Then we can see the pattern more clearly and practice alternative thoughts.
\end{quote}

\textbf{Interpretation.}
This package does not merely set a goal or assign homework in isolation. It compresses goal negotiation and structured follow-up monitoring into a single executable loop.
\end{casebox}

\vspace{0.8em}

\begin{casebox}{CBT: Rated Reappraisal Loop}
\small
\begin{tabularx}{\linewidth}{@{}p{0.22\linewidth}X@{}}
\textbf{Constituent micro-skills} &
Generate and evaluate alternative thoughts; rate belief and emotion before and after reappraisal. \\
\textbf{Representative functional skill} &
\texttt{Rated Reappraisal Loop}. \\
\textbf{Frequency} &
Internalization: 22; SFT: 104. \\
\end{tabularx}

\vspace{0.4em}
\textbf{Dialogue evidence.}
\begin{quote}\small
\textbf{Client:} The worst case is that everyone will remember me making a fool of myself and quietly laugh at me in class. The most likely case, though, is what we discussed before: the others were busy with their own practice, and the teacher just came over to help. \\
\textbf{Counselor:} Based on that evidence, let us replace the old thought with something more realistic and helpful: ``Yoga is a process of learning and caring for my body; most people are focused on their own practice; I am allowed to make mistakes while learning.'' If you use this alternative thought, how much do you now believe the old thought, how strong is the fear now, and how helpful does the new thought feel?
\end{quote}

\textbf{Interpretation.}
The package completes reappraisal and immediate feedback in the same turn, yielding a compact micro-intervention rather than a loose coexistence of two CBT labels.
\end{casebox}

\vspace{0.8em}

\begin{casebox}{BT: Dose-Specified Practice Scaffold}
\small
\begin{tabularx}{\linewidth}{@{}p{0.22\linewidth}X@{}}
\textbf{Constituent micro-skills} &
Specify practice duration and frequency; assign and monitor daily relaxation or exposure-related homework. \\
\textbf{Representative functional skill} &
\texttt{Dose-Specified Practice Scaffold}. \\
\textbf{Frequency} &
Internalization: 18; SFT: 72. \\
\end{tabularx}

\vspace{0.4em}
\textbf{Dialogue evidence.}
\begin{quote}\small
\textbf{Client:} I want to try this first when closing the shop and when I wake up in the middle of the night. Yesterday, when closing the shop, I tried breathing before checking, and it did help. \\
\textbf{Counselor:} Let us make the parameters explicit this week: after each exposure, delay checking for 3--5 minutes; do not seek extra checking or reassurance; if anxiety keeps rising for more than a minute, first do two rounds of breathing and attention redirection; record the start, peak, and end ratings, the urge for safety behaviors, the number of checks, and the learning point of that trial.
\end{quote}

\textbf{Interpretation.}
This package turns generic BT practice into a structured adherence scaffold with thresholds, recording rules, and a built-in procedure for continuation versus pause.
\end{casebox}

\vspace{0.8em}

\begin{casebox}{PDT: Validation-Wrapped Pacing Regulation}
\small
\begin{tabularx}{\linewidth}{@{}p{0.22\linewidth}X@{}}
\textbf{Constituent micro-skills} &
Normalize the pace of change and expectation; wrap interpretive pointing with empathy and validation. \\
\textbf{Representative functional skill} &
\texttt{Validation-Wrapped Pacing Regulation}. \\
\textbf{Frequency} &
Internalization: 6; SFT: 0. \\
\end{tabularx}

\vspace{0.4em}
\textbf{Dialogue evidence.}
\begin{quote}\small
\textbf{Client:} Every time I watch \emph{South Park}, I know I should stop after one episode and get back to work, but my finger just clicks the next one. Then I hate myself for having so little self-control. \\
\textbf{Counselor:} You have already been trying to set boundaries around rest, and that is not easy. Let us put ``Am I degenerating?'' under the category of self-evaluation, and put ``How long do I watch and when do I stop?'' under the category of entertainment. Would you be willing to speak to yourself using these two categories?
\end{quote}

\textbf{Interpretation.}
The package does more than combine validation and pacing. It first reduces shame and threat, then re-organizes the client’s experience into a more tolerable frame for further work.
\end{casebox}

\vspace{0.8em}

\begin{casebox}{PMT: Alternative-Story Amplification}
\small
\begin{tabularx}{\linewidth}{@{}p{0.22\linewidth}X@{}}
\textbf{Constituent micro-skills} &
Ask open questions linked to an alternative story; affirm abilities and highlight the meaning of change. \\
\textbf{Representative functional skill} &
\texttt{Alternative-Story Amplification}. \\
\textbf{Frequency} &
Internalization: 241; SFT: 1338. \\
\end{tabularx}

\vspace{0.4em}
\textbf{Dialogue evidence.}
\begin{quote}\small
\textbf{Client:} I want to write about the 7-Eleven owner giving me curry fish balls. Last Wednesday, when I walked in, he immediately said, ``The usual, right?'' and even gave me tissues. In that moment, I felt as if someone really remembered me. \\
\textbf{Counselor:} As a witness, I want to preserve the evidence you wrote into this new story today: the moments of being remembered, the way you stayed one minute longer in the tea room, and the family ritual you scheduled with concrete time and indicators. You have brought the ``taste of being remembered'' into daily life and expanded your support network. 
\end{quote}

\textbf{Interpretation.}
This package opens an alternative story and then immediately consolidates it through meaning and competence, producing a stable narrative-strengthening unit.
\end{casebox}

\vspace{0.8em}

\begin{casebox}{HET: Resource-to-Action Scaffold}
\small
\begin{tabularx}{\linewidth}{@{}p{0.22\linewidth}X@{}}
\textbf{Constituent micro-skills} &
Identify available resources and formulate near-term action goals; specify and implement an ongoing action plan. \\
\textbf{Representative functional skill} &
\texttt{Resource-to-Action Scaffold}. \\
\textbf{Frequency} &
Internalization: 54; SFT: 166. \\
\end{tabularx}

\vspace{0.4em}
\textbf{Dialogue evidence.}
\begin{quote}\small
\textbf{Client:} I choose to take three deep breaths before continuing the problem set. The rock in my chest feels a little lighter, but then I immediately think, ``This is such a waste of time.'' \\
\textbf{Counselor:} Today we continued practicing the distinction between facts, interpretations, and bodily signals, and we brought ``I choose / I am willing / I take responsibility'' into relationships and boundaries. We jointly set a one-week social exposure ladder and paired it with breathing, self-supportive language, and structured records of what happened, how you interpreted it, what your body said, what you chose, and what followed.
\end{quote}

\textbf{Interpretation.}
Resource identification does not remain abstract. It is immediately translated into a concrete action ladder, self-positioning language, and a monitoring routine, forming a complete practice scaffold.
\end{casebox}

Across therapies, these cases support the same conclusion: the model increasingly relies on recurrent functional packages built from existing skills. This pattern supports a conservative claim of \emph{skill-level compositional generalization beyond the original inventory}: the model is not primarily inventing entirely new techniques, but reorganizing known micro-skills into higher-order functional units with clearer operational value.

\subsection{Cross-Therapy Reuse}
\label{sec:cross_therapy_reuse}

A second important effect of internalization is that some of these functional packages become reusable across therapies. Here, however, the strongest evidence does not come from exact cross-therapy bundles within the same turn, since such strict mixed-therapy combinations remain limited. Instead, the more informative pattern is \emph{cross-therapy reuse at the functional-family level}: similar intervention logics recur across different therapy-specific contexts while preserving related local counseling goals.

\begin{table}[t]
\centering
\caption{Representative functional families reused across therapies after internalization. The main evidence comes from family-level reuse rather than from frequent exact mixed-therapy bundles in a single turn.}
\label{tab:cross_therapy_reuse}
\small
\setlength{\tabcolsep}{4pt}
\begin{tabular}{l l c c}
\toprule
\textbf{Functional family} & \textbf{Therapies involved} & \textbf{Internal} & \textbf{SFT} \\
\midrule
Progress Consolidation Package & BT, HET, PDT, PMT & 292 & 2287 \\
Goal-to-Practice Bridge & BT, CBT, HET, PMT & 207 & 984 \\
Affect Attunement Before Deeper Work & CBT, HET, PDT & 28 & 48 \\
\bottomrule
\end{tabular}
\end{table}

\begin{casebox}{Cross-Therapy Reuse: Goal-to-Practice Bridge in a CBT Context}
\small
\begin{tabularx}{\linewidth}{@{}p{0.22\linewidth}X@{}}
\textbf{Functional family} &
\texttt{Goal-to-Practice Bridge}. \\
\textbf{Constituent micro-skills} &
Negotiate and clarify near-term goals; assign between-session practice and adjust it based on feedback. \\
\textbf{Therapies involved} &
BT, CBT, HET, PMT. \\
\textbf{Frequency} &
Internalization: 207; SFT: 984. \\
\end{tabularx}

\vspace{0.4em}
\textbf{Dialogue evidence.}
\begin{quote}\small
\textbf{Client:} Honestly, I do not think writing these things down will help much. Yesterday I saw a couple again, felt really heavy inside, and just turned around and went home. \\
\textbf{Counselor:} Today we identified the chain from trigger to thought to intense emotion to avoidance, noticed several cognitive patterns, and agreed on a one-week small experiment with brief records, plus a ``pause, breathe, then observe again'' step. Next week we will review the records and adjust the goals and methods together.
\end{quote}

\textbf{Interpretation.}
What recurs here is not merely a skill label, but the same functional principle: insight is immediately translated into a concrete, reviewable practice plan.
\end{casebox}

\vspace{0.8em}

\begin{casebox}{Cross-Therapy Reuse: Progress Consolidation Package}
\small
\begin{tabularx}{\linewidth}{@{}p{0.22\linewidth}X@{}}
\textbf{Functional family} &
\texttt{Progress Consolidation Package}. \\
\textbf{Constituent micro-skills} &
Provide concrete affirmation of progress and strengths; systematically review key strategies and growth. \\
\textbf{Therapies involved} &
BT, HET, PDT, PMT. \\
\textbf{Frequency} &
Internalization: 292; SFT: 2287. \\
\end{tabularx}

\vspace{0.4em}
\textbf{Dialogue evidence.}
\begin{quote}\small
\textbf{Client:} The ``pause for a second and look at the brow'' trick helped the most. At a client dinner, I tried it and did not rush to fill the silence like before. \\
\textbf{Counselor:} You have turned ``pause -- look for evidence -- breathe slowly -- delay inference'' into your own stability switch. Let us define it as your emergency kit, pair it with brief weekly rehearsal and structured records, and treat the wider interval before the next session not as abandonment, but as returning agency to you.
\end{quote}

\textbf{Interpretation.}
The counselor does not merely praise improvement. Progress is summarized, named as a replicable capacity, and converted into a maintainable future direction.
\end{casebox}

\vspace{0.8em}

\begin{casebox}{Cross-Therapy Reuse: Progress Consolidation Package (Another Recurrence)}
\small
\begin{tabularx}{\linewidth}{@{}p{0.22\linewidth}X@{}}
\textbf{Functional family} &
\texttt{Progress Consolidation Package}. \\
\textbf{Constituent micro-skills} &
Provide concrete encouragement; review homework with positive reinforcement and suggestions for refinement. \\
\textbf{Therapy of this case} &
BT. \\
\end{tabularx}

\vspace{0.4em}
\textbf{Dialogue evidence.}
\begin{quote}\small
\textbf{Client:} I will try saying it this way: when I really want to tear up the drawing, I first wait five minutes, breathe twice, look at my cue card, and write down three small things I can improve. But I still am not sure I can really do it. \\
\textbf{Counselor:} You explained the steps clearly and also told me honestly about your worry. Let us turn the delay into a two-stage process, pair it with breathing, the cue card, and small rewards, and bring back written records next time. You have already made steady attempts, and now we are making them into a reproducible process.
\end{quote}

\textbf{Interpretation.}
This recurrence strengthens the claim that consolidation here is a reusable functional family rather than a one-off wording choice.
\end{casebox}

\vspace{0.8em}

\begin{casebox}{Cross-Therapy Reuse: Affect Attunement Before Deeper Work}
\small
\begin{tabularx}{\linewidth}{@{}p{0.22\linewidth}X@{}}
\textbf{Functional family} &
\texttt{Affect Attunement Before Deeper Work}. \\
\textbf{Constituent micro-skills} &
Identify emotion before probing automatic thought; mark the linked emotional and behavioral response. \\
\textbf{Therapies involved} &
CBT, HET, PDT. \\
\textbf{Frequency} &
Internalization: 28; SFT: 48. \\
\end{tabularx}

\vspace{0.4em}
\textbf{Dialogue evidence.}
\begin{quote}\small
\textbf{Client:} It was Wednesday night. The message showed as read but there was no reply. I immediately thought, ``Of course this happens again -- who would ever be interested in someone like me?'' It hurt at about an eight out of ten. \\
\textbf{Counselor:} Thank you for describing that so concretely. Could you choose another record this week where you used the ``pause and breathe/hot drink'' step, and tell it in the order of time and situation, thought, emotion rating, what you did next, and how you felt afterward?
\end{quote}

\textbf{Interpretation.}
What is reused here is the structure of first stabilizing and naming the emotional experience, and only then moving toward deeper cognitive or experiential work.
\end{casebox}

Several reusable functional families illustrate the same pattern. \emph{Goal-to-Practice Bridge} appears across multiple therapies as a shared logic for translating insight, discussion, or formulation into concrete between-session action. \emph{Progress Consolidation Package} recurs as a stable way of summarizing gains, affirming client competence, and orienting the next phase of work. \emph{Affect Attunement Before Deeper Work} shows another transferable structure, in which the counselor first stabilizes or names the client’s emotional experience before moving into more challenging exploratory work. In each case, what travels across therapies is not merely a skill label, but an intervention principle.

This distinction is important. The evidence does not support the stronger claim that internalization turns the model into a fully therapy-agnostic counselor, nor that it frequently produces large numbers of exact cross-therapy bundles. What it does support is a more careful conclusion: internalization enables some higher-order functional skills to be reused in multiple therapeutic contexts, suggesting that the model learns counseling principles that are portable beyond a single therapy-specific template.

Taken together, these findings indicate that internalization improves both the compositional organization and the portability of counseling skills. The model becomes better at turning existing micro-skills into reusable functional units, and some of these units begin to generalize across therapies. In this sense, internalization moves the model toward a more general and versatile counselor.

\end{document}